\documentclass[10pt,twocolumn,letterpaper]{article}

\usepackage{cvpr}              
\usepackage[T1]{fontenc}
\usepackage{threeparttable}
\usepackage{makecell}

\usepackage{float}
\usepackage{algorithm} 
\usepackage{makecell}
\usepackage{multirow}
\usepackage{algpseudocode}
\usepackage{booktabs}
\usepackage{amssymb}

\definecolor{cvprblue}{rgb}{0.21,0.49,0.74}
\usepackage[pagebackref,breaklinks,colorlinks,allcolors=cvprblue]{hyperref}

\def\paperID{10543} 
\def\confName{CVPR}
\def\confYear{2026}

\title{~\includegraphics[height=15pt]{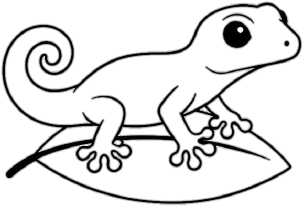}
GeCo-SRT: Geometry-aware Continual Adaptation for

Robotic Cross-Task Sim-to-Real Transfer
}

\author{
Wenbo Yu$^{1,*}$ \quad Wenke Xia$^{2,3,4,}$\thanks{Equal contribution.} \quad Weitao Zhang$^{2}$ \quad Di Hu$^{2,3,4,}$\thanks{Corresponding author.} \\
$^{1}$Beijing Forestry University, Beijing, China \\
$^{2}$Gaoling School of Artificial Intelligence, Renmin University of China, Beijing, China\\
$^{3}$Beijing Key Laboratory of Research on Large Models and Intelligent Governance\\
$^{4}$Engineering Research Center of Next-Generation Intelligent Search and Recommendation, MOE\\
}
\begin{document}
\maketitle
\begin{abstract}

Bridging the sim-to-real gap is important for applying low-cost simulation data to real-world robotic systems. However, previous methods are severely limited by treating each transfer as an isolated endeavor, demanding repeated, costly tuning and wasting prior transfer experience.
To move beyond isolated sim-to-real, we build a continual cross-task sim-to-real transfer paradigm centered on \textbf{knowledge accumulation across iterative transfers, thereby enabling effective and efficient adaptation to novel tasks.} 
Thus, we propose GeCo-SRT, a geometry-aware continual adaptation method. It utilizes domain-invariant and task-invariant knowledge from local geometric features as a transferable foundation to accelerate adaptation during subsequent sim-to-real transfers. 
This method starts with a geometry-aware mixture-of-experts module, which dynamically activates experts to specialize in distinct geometric knowledge to bridge observation sim-to-real gap. Further, the geometry-expert-guided prioritized experience replay module preferentially samples from underutilized experts, refreshing specialized knowledge to combat forgetting and maintain robust cross-task performance.
Leveraging knowledge accumulated during iterative transfer, GeCo-SRT method not only achieves 52\% average performance improvement over the baseline, but also demonstrates significant data efficiency for new task adaptation with only 1/6 data.
We hope this work inspires approaches for efficient, low-cost cross-task sim-to-real transfer.

\end{abstract}    
\section{Introduction}
\label{sec:intro}

Leveraging scalable and low-cost simulation data to train real-world robotic systems offers an ideal pathway to significantly accelerate robot learning and deployment~\cite{han2025wheeledlabmodernsim2real,9308468,da2025surveysimtorealmethodsrl,james2019simtorealsimtosimdataefficientrobotic,peng2018sim}. However, the inevitable discrepancies between the simulated model and physical reality, such as unrealistic visual rendering and simplified physical dynamics~\cite{jeong2019selfsupervisedsimtorealadaptationvisual,torne2024reconciling,feng2025learning,lu2026visionproprioceptionpoliciesfailrobotic}, cause policies learned in simulation to suffer significant performance degradation upon real-world deployment~\cite{aljalbout2025realitygaproboticschallenges,DBLP:journals/corr/abs-2009-13303}. This phenomenon, known as the ``sim-to-real gap'', presents a critical and long-standing challenge that impedes the acceleration of real-world robot learning through simulation~\cite{DBLP:journals/corr/abs-2009-13303,9308468}.

\begin{figure}[t]
  \centering  
  \includegraphics[width=\linewidth]{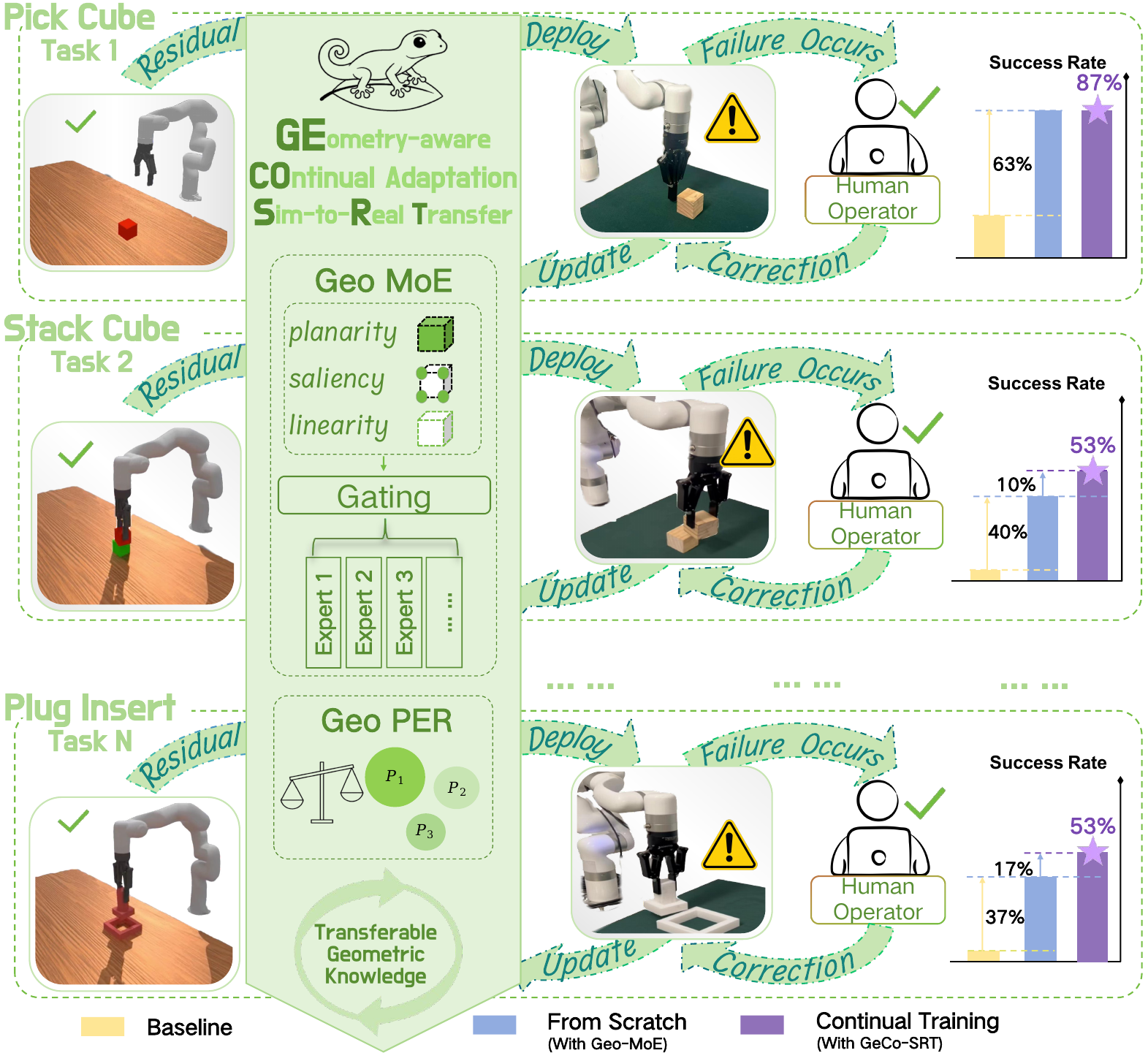}  
  \caption{
 Our GeCo-SRT method accumulates transferable geometric knowledge from sequential sim-to-real tasks (Task $1...N$), which facilitates effective and efficient transfer to novel tasks.
  }
  \label{fig:teaser}  
  \vspace{-1em}
\end{figure}
Several strategies have been proposed to address the sim-to-real gap~\cite{DBLP:journals/corr/abs-2009-13303,he2025asapaligningsimulationrealworld,ma2023sim2real2activelybuildingexplicit}, but they often face limitations in efficiency and generalization. For instance, System Identification seeks to model real-world physics but is labor-intensive and struggles with complex dynamics~\cite{memmel2024asidactiveexplorationidentification,sobanbabu2025samplingbasedidentificationactiveexploration}. While Domain Randomization offers an alternative by training policies across diverse simulations~\cite{mehta2019activedomainrandomization,Volpi_2021_CVPR}, its effectiveness hinges on manually-configured randomization ranges that may not fully capture real-world properties. Although recent data-driven transfer methods~\cite{huo2022domainagnosticpriortransfersemantic,8416737} mitigate the reliance on manual design, they remain limited to single task adaptation.
This highlights a common drawback across existing approaches: they treat each sim-to-real transfer as an isolated process, which inevitably demands repeated and costly efforts for new task adaptation. 
This limitation points to a critical but under-explored challenge: how to \textit{leverage past experiences to facilitate effective and efficient sim-to-real transfer on new tasks}, rather than requiring isolated, task-specific tuning from scratch for each.

To move beyond the current paradigm of isolated sim-to-real transfer, we propose a new continual cross-task sim-to-real transfer paradigm. As shown in Figure~\ref{fig:teaser}, this paradigm is designed to achieve continual knowledge accumulation across iterative transfers from previous experiences (e.g., ``pick cube''), thereby enabling effective and efficient adaptation to novel tasks (e.g., ``stack cube''). This process hinges on a robust medium for transferable knowledge that bridges both tasks and domains. To this end, we leverage local geometric features as the primary source of transferable knowledge due to their crucial dual invariance. On the one hand, local geometric features are domain-invariant,
offering a robust solution to the sim-to-real gap. Different from visual perception, which is highly sensitive to domain shifts in texture and material properties, local geometric features (e.g., surface normals) derived from 3D data are structurally consistent across simulation and reality. On the other hand, it is also task-invariant, as these primitives are shared across diverse manipulation tasks. For example, the same flat surface of the cube that informs a grasping pose for the ``pick cube'' task is also the critical target feature for the ``stack cube'' task. This dual invariance makes local geometric features a robust knowledge medium, enabling effective generalization and continual knowledge accumulation.

Building on this premise, we introduce the GeCo-SRT method, which is a geometry-aware continual adaptation method to achieve continual knowledge accumulation across iterative sim-to-real transfers. 
Concretely, our GeCo-SRT method employs the human-in-the-loop method~\cite{jiang2024transicsimtorealpolicytransfer} to reframe sim-to-real transfer as learning from human correction trajectories. This process quantifies the sim-to-real gap into a continuously growing replay buffer, enabling knowledge accumulation across tasks. Further, we propose the Geometry-aware Mixture-of-Experts (Geo-MoE) module, which serves as a reusable perception residual network for point cloud to bridge the observation gap. This module first extracts local geometric features from the input point cloud and utilizes these local geometric features as cues to dynamically activate experts, thereby achieving distinct knowledge reuse across different tasks.
Furthermore, to protect this specialized geometric knowledge within the experts during continual learning, we introduce Geometry-expert-guided Prioritized Experience Replay (Geo-PER) module. This strategy combats catastrophic forgetting by dynamically prioritizing historical samples based on expert utility, ensuring the model retains its ability to process key geometric structures from previous tasks while adapting to new ones. 
Through our GeCo-SRT method, we achieve effective and efficient sim-to-real transfer across various tasks by continually leveraging geometric knowledge extracted from prior sim-to-real deployments.

In this work, we establish a sequence of 4 sim-to-real robotic manipulation tasks to evaluate continual sim-to-real transfer. The results of single task and continual cross-task sim-to-real transfer experiments prove the effectiveness and efficiency of our geometry-aware continual adaptation method. Specifically, our method ultimately outperforms the baseline by an average of 52\% while demonstrating remarkable data efficiency, matching the baseline's success rate with only 16.7\% of the data. This validates that our geometry-aware method can effectively extract reusable knowledge for continual sim-to-real transfer. We hope this work will inspire future research on effective and efficient cross-task sim-to-real transfer method exploration.

\section{Related Work}
\label{sec:related}
\subsection{Robot Learning via Sim-to-Real Transfer}
Bridging the gap between simulation and reality remains a fundamental and extensively studied challenge in robot learning~\cite{mu2021maniskillgeneralizablemanipulationskill,9606868,arndt2019metareinforcementlearningsimtoreal,bousmalis2017usingsimulationdomainadaptation}.
While System Identification (SysID) seeks to model the physical parameters of the real system, it relies on pre-defined model structures and extensive real-world data collection for parameter tuning~\cite{memmel2024asidactiveexplorationidentification}. However, this process is highly labor-intensive and proves inadequate for modeling complex, non-linear dynamics. 
Alternatively, Domain Randomization trains policy across a wide distribution of simulated variations, aiming to treat the real world as another variant~\cite{mehta2019activedomainrandomization,chen2022understandingdomainrandomizationsimtoreal}. This approach requires expert-driven tuning of the randomization ranges, which may fail to effectively cover the real-world system dynamics.
Alternatively, Transic~\cite{jiang2024transicsimtorealpolicytransfer} employs online correction trajectories to achieve domain-agnostic sim-to-real transfer without explicit modeling. 
However, these approaches are limited to single task sim-to-real transfer, which makes the adaptation of each new task from simulation to the real world a costly endeavor.
In this work, we propose the geometry-aware continual adaptation method to accumulate geometric knowledge from previous sim-to-real transfer for efficient novel task adaptation.

\begin{figure*}[t]
  \vspace{-1em}
  \centering  
  \includegraphics[width=1.0\linewidth]{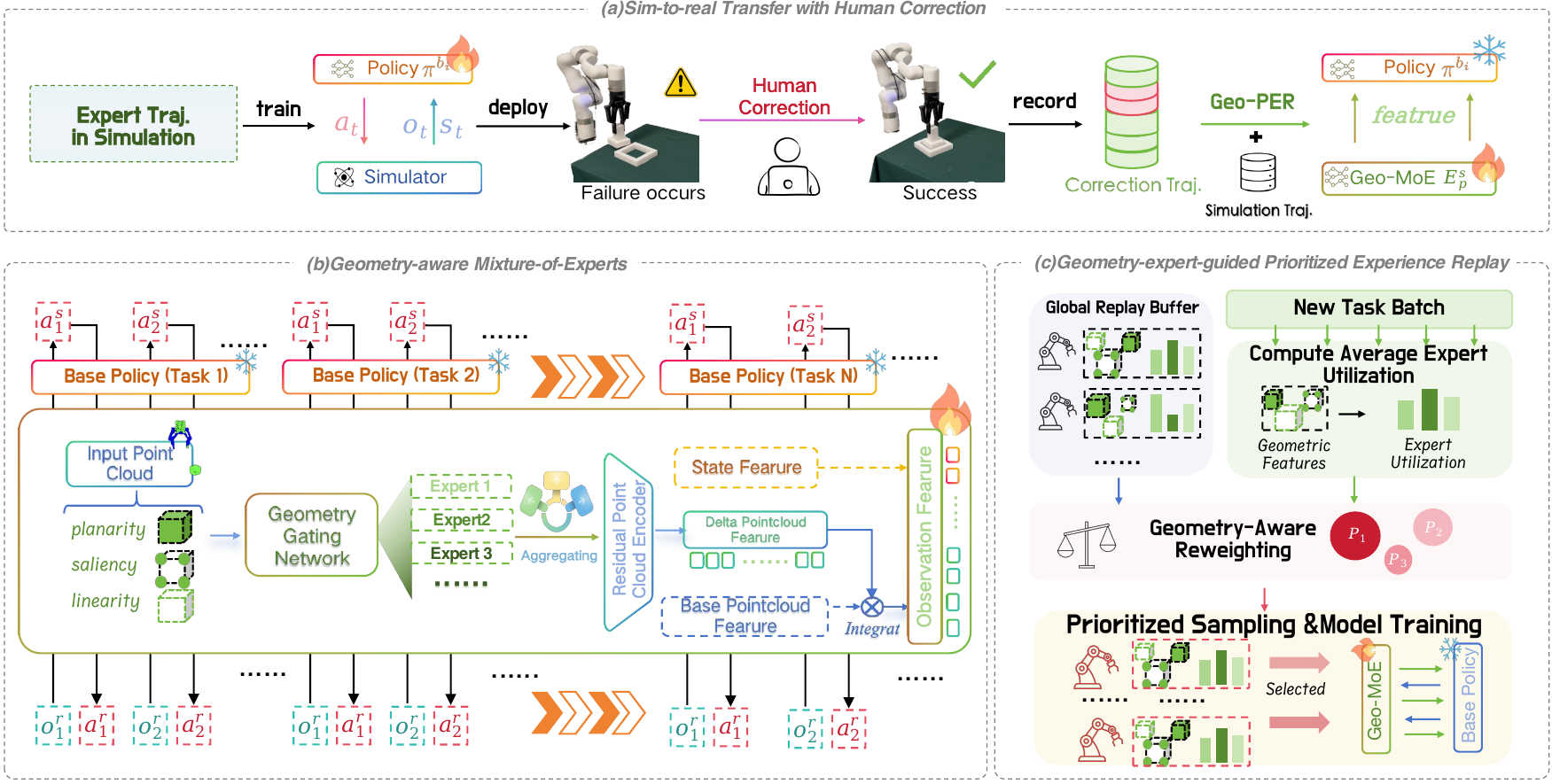}  
  \caption{Overview of our geometry-aware continual adaptation method. (a) demonstrates the sim-to-real transfer with human correction pipeline, which translates sim-to-real heuristics into learnable human correction trajectories. (b) represents the Geometry-aware Mixture-of-Experts module, which extracts local geometric features to dynamically activate experts, enabling them to specialize in distinct geometric knowledge for effective sim-to-real transfer. (c) illustrates the Geometry-expert-guided Prioritized Experience Replay, which leverages expert utilization to prioritize historical data, thereby mitigating catastrophic forgetting and promoting knowledge reuse.
  }
  \label{fig:pipeline}  
  \vspace{-1em}
\end{figure*}

\subsection{Continual Learning in Robotic Manipulation}
Continual learning, known as the ability to learn new skills sequentially without forgetting previous ones, is essential for autonomous robots adapting to a dynamic world. To mitigate catastrophic forgetting, replay-based methods~\cite{smith2021dreamingnewapproachdatafree,DBLP:journals/corr/abs-2009-13303,Smith_2021_ICCV} store a subset of data from past tasks in memory buffer and interleave this data when training on new tasks. Since this replay buffer can be memory-intensive, other works propose using a generative model~\cite{heppert2025realgen,10610566} to produce pseudo-samples from past data distributions, thereby reducing memory overhead~\cite{xia2023flashllmenablingcosteffectivehighlyefficient,wang2023serialcontrastiveknowledgedistillation,xia2025phoenix}. Alternatively, regularization-based methods~\cite{Jung_2025,knoedler2022improvingpedestrianpredictionmodels} aim to protect knowledge by constraining updates to network parameters deemed important for previous tasks.
Although these methods have proven effective in classification tasks, the characteristics of robotic manipulation, such as high-dimensional continuous state-action spaces and long-horizon temporal dependencies, make the direct application of these methods challenging. Thus, recent efforts have focused on developing dedicated benchmarks, such as LIBERO~\cite{liu2023liberobenchmarkingknowledgetransfer} for assessing knowledge transfer. 
Concurrently, novel algorithmic paradigms~\cite{wan2024lotuscontinualimitationlearning,powers2023evaluatingcontinuallearninghome,zeng2024mpi,xia2025robotic} have been proposed to specifically address the challenges of robotic manipulation.
However, these works only focus on continual learning for robotic imitation learning paradigm. In this work, we introduce continual learning into cross-task sim-to-real transfer, 
and propose the geometry-expert-guided prioritized experience replay module to utilize geometric knowledge to reweight replay samples for novel task sim-to-real transfer while simultaneously alleviating catastrophic forgetting.

\section{Geometry-aware Continual Adaptation}
The persistent sim-to-real gap presents a significant challenge for robotic policy sim-to-real transfer. 
While prior work often addresses sim-to-real transfer as an individual, isolated process, we propose the GeCo-SRT, a geometry-aware continual adaptation method, which is designed to continually accumulate knowledge across iterative sim-to-real transfers for effective and efficient adaptation to novel tasks. To achieve this, our method is designed to address two fundamental questions:
\begin{itemize}
    \item How to convert human-designed sim-to-real heuristics into a learnable format for neural networks?
    \item What specific knowledge is transferable across the iterative sim-to-real transfer process?
\end{itemize}

\par To address the first problem, our GeCo-SRT method proposes a human-in-the-loop method to convert sim-to-real transfer as learning from human correction trajectories, making the process suitable for iterative continual learning as shown in Section~\ref{sec:base_simulation}.
\par In response to the second question, we posit that local geometric features serve as the core transferable knowledge. We therefore propose the Geometry-aware Mixture-of-Experts (Geo-MoE) to extract and reuse these features for efficient cross-task adaptation as shown in Section~\ref{sec:geo_moe}. Finally, we introduce the Geometry-expert-guided Prioritized Experience Replay (Geo-PER) in Section~\ref{sec:geo_per} to protect this specialized knowledge by dynamically prioritizing historical samples, ensuring its retention during continual adaptation.
The method is shown in Figure~\ref{fig:pipeline}.

\subsection{Sim-to-real Transfer with Human Correction}\label{sec:base_simulation}
For each task $i$, we train a base policy $\pi^{b_i}$ in simulation using expert trajectories with 3D point cloud observations. This policy consists
 of a point cloud encoder $E_p^{b_i}$ and a diffusion policy head $\pi_h^{b_i}$. Further, we deploy the base policy to real-world environment and collect human corrections dataset $\mathcal{D}^{h_i}_{real}$ via a shared-autonomy framework. 
 Finally, to bridge the sim-to-real gap, we introduce a shared perception residual module $E_p^s$. During the sim-to-real transfer, this shared module is continually updated using a hybrid replay buffer that draws from both the simulation dataset and the newly collected real-world correction dataset $\mathcal{D}^{h_i}_{real}$.

\subsubsection{Base Policy Training in Simulation}
To bypass the drawbacks of RGB (e.g., vulnerability to camera poses as well as discrepancies between synthetic and real images), we use 3D point clouds as the primary visual modality to better facilitate sim-to-real transfer.

For task $i$, we generate an expert dataset $D_{sim}^{e_i}$ using a motion planner in simulation. For observations, we capture depth images from two cameras, which are then transformed and aggregated into a complete scene point cloud $P$ in the robot's base frame. This process mirrors our real-robot setup to minimize the sim-to-real gap. The resulting $P$ is then downsampled via Farthest Point Sampling to serve as the policy network input.

We train our base diffusion policy $\pi^{b_i}$, using behavior cloning on the expert dataset $D_{sim}^{e_i}$. This policy is parameterized as the noise prediction network $\epsilon_{\theta}$. It is optimized using the simplified L2 diffusion loss:
\begin{equation}
\mathcal{L}_{\text{diff}} = \mathbb{E}_{ (a_t,o_t) \sim \mathcal{D}^{e_i}_{sim}, \epsilon \sim \mathcal{N}(0, I)} \left[ \left\| \epsilon - \epsilon_{\theta}(a_t^k, k, o_t) \right\|^2 \right]
\end{equation}
where $o_t$ includes the proprioceptive state $s_t$ and point cloud $P$, $a_t$ is the corresponding expert action.

\subsubsection{Continual Learning with Human Correction}
For each task $i$, we deploy the base policy $\pi^{b_i}$ in the real world under human supervision via a shared-autonomy framework. 
At each timestep $t$, if the operator anticipates a failure, they intervene with a teleoperated action $a_t \leftarrow a_t^h$, setting an indicator $I_t^h$ to true. Otherwise, the base policy acts, $a_t \leftarrow \pi^{b_i}(o_t)$, and $I_t^h$ is set to false. We collect these tuples $(o_t, a_t, I_t^h)$ to form the real-world human correction dataset $\mathcal{D}^{h_i}_{real}$.
Further, we merge the correction dataset $\mathcal{D}^{h_i}_{real}$ with the simulation trajectories $D_{sim}^{e_i}$ to form the replay buffer $D_{buf}^{m_i}$. This buffer is used to train the perception residual module $E_{p}^{s}$, which aligns the point cloud observation feature to bridge the sim-to-real gap.

As shown in Figure~\ref{fig:pipeline}(a), while the parameters of the base simulation policy $\pi^{b_i}$ are frozen, the perception residual module $E_p^s$ is maintained as a shared module across all tasks. This allows it to progressively learn and consolidate reusable knowledge from the buffer $D_{buf}^{m_i}$ of each task, enabling effective iterative sim-to-real transfer.

\subsection{Geometry-Aware Mixture-of-Experts}\label{sec:geo_moe}

We posit that local geometric information is the ideal primary source of transferable knowledge for continual sim-to-real transfer due to the crucial dual invariance:

\begin{itemize}
    \item Domain Invariance: Unlike textures or dynamics, local geometry (e.g., planarity and linearity) remains consistent between simulation and reality, providing a stable bridge for the sim-to-real gap.
    \item Task Invariance: Geometric primitives (e.g., edges, corners) are shared across distinct manipulation tasks (e.g., ``pick cube'' vs. ``stack cube''), enabling the continual accumulation of knowledge.
\end{itemize}

Based on this premise, we propose the Geometry-aware Mixture-of-Experts (Geo-MoE) as our perception residual module $E_p^s$. It processes local geometric discrepancies to dynamically activate specialized experts, thereby adapting the feature representation and enabling the reuse of geometric knowledge across tasks.

Concretely, our Geo-MoE module first samples local point groups $g_i$ from the input point cloud $P$ using the k-Nearest Neighbors (k-NN) method~\cite{10.1007/978-3-540-39964-3_62}. For each group $g_i$, we estimate its local geometric features—planarity, linearity, and saliency—using the local PCA calculation method~\cite{DBLP:journals/corr/abs-2111-12663}. The group $g_i$ is then processed by a lightweight gating network $G$ to produce routing weights $w_i = \text{Softmax}(G(g_i))$ for $M$ parallel expert networks. The final processed feature for each group is the weighted sum:
\begin{equation}
g'_i = \sum_{j=1}^{M} w_{i,j} \cdot \text{Expert}_j(g_i).
\end{equation}
These per-group features $\{g'_i\}$ are ultimately aggregated to form the single corrective residual vector $g'_{res}$. This observation residual, $g'_{res}$, is then concatenated with the output of the frozen base encoder $E_p^{b_i}$ to form a corrected feature vector: $\hat{f} = \text{Concat}(E_p^{b_i}(o), g'_{res})$.
Finally, this corrected feature $\hat{f}$ is passed to the diffusion policy head $\pi_h^{b_i}$ to produce the final action $\hat{a} = \pi_h^{b_i}(\hat{f})$.

During training, we keep the base policy $\pi^{b_i}$ frozen and exclusively update the Geo-MoE module $E_p^s$ using the mixed replay buffer $D_{buf}^{m_i}$:
\begin{equation}
L_{total} = MSE(\hat{a},a) + \alpha L_{balance},
\end{equation}
where $L_{balance}$ is a standard auxiliary loss~\cite{wang2024auxiliarylossfreeloadbalancingstrategy,pmlr-v162-rajbhandari22a} to ensure balanced expert utilization and prevent gating collapse.

\subsection{Geometry-Expert-Guided Prioritized 
Experience Replay} \label{sec:geo_per}

In our continual sim-to-real setting, the model adapts sequentially across $N$ tasks. After $N-1$ tasks, the data is stored in a unified replay buffer $\mathcal{R}$, the union of all preceding per-task mixed buffers:
$\mathcal{R} = \bigcup_{j=1}^{N-1} D_{buf}^{m_j}$.

When adapting to the new N-th task, the key challenge is preventing expert-level catastrophic forgetting. Standard Prioritized Experience Replay (PER) methods, which prioritize samples based on task loss, are ill-suited for this. By ignoring expert-specific utility, they neglect idle experts, whose specialized knowledge is subsequently forgotten.
To address this deficiency, we introduce Geometry-expert-guided Prioritized Experience Replay (Geo-PER). Our strategy fundamentally shifts the prioritization metric: from task loss to expert utilization. This ensures all experts are periodically refreshed with relevant data from R, even when inactive in the current task N.

Our Geo-PER mechanism operates as follows: First, we store the expert activation vector $W_i = \{w_{i,1}, \dots, w_{i,M}\}$ associated with each historical sample $i \in \mathcal{R}$. When processing new data from the current task $N$, we compute the average expert utilization $U^{\text{new}} = \{u_1^{\text{new}}, \dots, u_M^{\text{new}}\}$. We then dynamically update the sampling priority $P_i$ for all historical data $i \in \mathcal{R}$:
\begin{equation}P_i \propto \sum_{j=1}^{M} w_{i,j} \cdot \frac{1}{u_j^{\text{new}} + \epsilon},
\end{equation}
where $w_{i,j}$ is the stored activation of expert $j$ for sample $i$, $u_j^{\text{new}}$ is the current task's average utilization of expert $j$, and $\epsilon$ is a stability constant. 
This prioritization scheme functions as an active counter-balance to expert imbalance. If an expert $j$ is under-utilized by the new task ($u_j^{\text{new}}$ is low), its reciprocal term $(u_j^{\text{new}} + \epsilon)^{-1}$ becomes large. Consequently, Geo-PER assigns a high sampling probability to historical samples (from tasks $1..N-1$) that strongly activated that expert ($w_{i,j}$ is high). This ensures that the parameters of idle experts are continuously included in gradient updates, thereby mitigating catastrophic forgetting.

\begin{figure*}[t]  
  \vspace{-1em}
  \centering  
  \includegraphics[width=\linewidth]{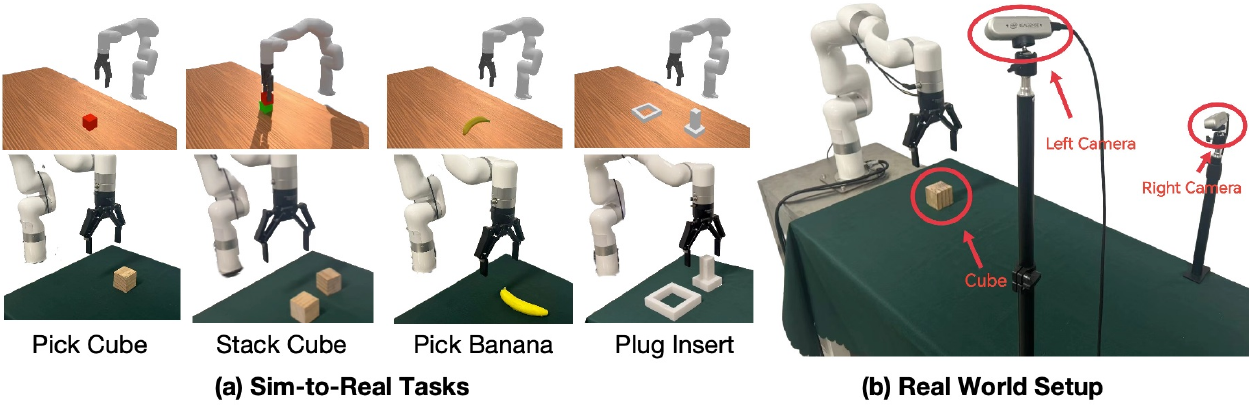}  
  \caption{We demonstrate the task setup and real-world setup. (a) illustrates 4 robotic manipulation task in simulation and real-world scenarios. (b) shows that the real-world setup, which consists of a XARM equipped with a robotiq-140 gripper.} 
  \label{fig:setup}  
  \vspace{-1em}
\end{figure*}

\section{Experiments}
\label{sec:experiment}
To comprehensively evaluate our method, we propose experiments to answer the following questions:

\begin{itemize}
    \item \textbf{Q1:} Could our geometry-aware method promote efficient continual sim-to-real transfer across different tasks? (Section~\ref{sec:comparison})
    \item \textbf{Q2:} What are the specific contributions of our Geo-MoE and Geo-PER method on continual cross-task sim-to-real transfer? (Section~\ref{sec:ablation})
    \item \textbf{Q3:} Would the task similarity influence the efficiency of knowledge transfer when adapting between different sim-to-real tasks? (Section~\ref{sec:cross_transfer})
\end{itemize}

\subsection{Implementation Details}
~\label{sec:details}
For empirical validation, we utilize four robotic manipulation tasks, spanning both the ManiSkill simulation environment and the real world (Fig.~\ref{fig:setup} (a)): \textit{Pick Cube}, \textit{Stack Cube}, \textit{Pick Banana}, and \textit{Plug Insert}. 

Our sim-to-real pipeline starts by training an initial policy in simulation using imitation learning on 2,000 expert trajectories. We then deploy this policy in the real world and gather 60 human correction trajectories. Finally, to ensure effective transfer and mitigate catastrophic forgetting, we train the final policy on a mixed dataset combining these new corrections with a replay buffer from prior tasks.

\subsubsection{Real-world Setup}

\par Our hardware (Fig.~\ref{fig:setup} (b)) comprises an Xarm5 arm and a Rotiq2F140 gripper. We utilize a RealSense D435 and a RealSense D435i to obtain the point cloud observation; their point clouds are transformed to the robot's base frame, fused, and cropped within a bounding box to remove background. Human correction trajectories are collected by operators providing real-time interventions via a 3Dconnexion SpaceMouse during execution.

\subsubsection{Training Details}
Our training involves two stages: simulation policy training and continual sim-to-real transfer.

In the simulation phase, we first train the diffusion policy with 3D point cloud encoder as the base simulation policy. The main hyperparameters for this stage were set as follows: learning rate of $3 \times 10^{-4}$, denoise steps are 10, and action chunking size is 8.

During the continual sim-to-real transfer, we employ a continual learning approach to acquire generalizable geometric knowledge to continually bridge the sim-to-real gap.  To retain knowledge from previous tasks, we set the mixing in 10\% of data from prior tasks (including both simulation and real-world trajectories). Human-corrected data constitutes 95\% of the training data in this stage.
In this process, the residual learning rate is set to $1 \times 10^{-3}$.

For the model architecture and experience replay strategy, each expert within our Geo-MoE module is trained with learning rate of $1 \times 10^{-3}$. For our Geo-PER strategy, the parameter controlling the degree of prioritization in sampling is set to 0.6. Additionally, a smoothing term to ensure non-zero sampling probability is set to $1 \times 10^{-6}$, and the EMA coefficient for priority updates is set to 0.4.

\subsubsection{Evaluation Metrics}

To comprehensively evaluate the performance of our continual cross-task sim-to-real approach, we employ two key metrics: Success Rate (SR) and Normalized Negative Backward Transfer (N-NBT). SR measures the cross-task forward transfer ability, while N-NBT quantifies the degree of catastrophic forgetting. For each task, we evaluate methods with 30 trials and report the average SR and N-NBT.

Normalized Negative Backward Transfer: Negative Backward Transfer (NBT)~\cite{liu2023libero} fails to accurately reflect forgetting when the initial success rate is low, as is common in sim-to-real. We therefore propose Normalized Negative Backward Transfer (N-NBT) to evaluate the \emph{relative} performance drop.
\begin{equation}
\text{N-NBT}_k = \frac{1}{K-k} \sum_{\tau=k+1}^K \left( \frac{C_{\tau,k} - C_{k,k}}{C_{k,k}} \right).
\end{equation}
A lower N-NBT value indicates better knowledge retention from prior tasks and thus less catastrophic forgetting.

\subsection{Comparison Results} \label{sec:comparison}
\par\par To evaluate our proposed Geometry-guided Continual Sim-to-real Transfer method, we conduct comprehensive comparisons against several baseline approaches. The evaluation is performed on two primary fronts: (1) single task sim-to-real transfer and (2) cross-task continual learning.

\subsubsection{Single Task Sim-to-Real Transfer}

\par We first evaluate single task adaptation against the following three baselines:

\begin{itemize}
    \item \textbf{Direct Deploy:} The simulation base policy directly deploys in the real-world environment without fine-tuning. 

    \item \textbf{Action Residual:} The action residual module is a residual diffusion policy head network, which takes the observation feature from frozen point cloud encoder and return the residual action. It serves as a baseline to quantify the sim-to-real gap through only action level. 

    \item \textbf{Transic~\cite{jiang2024transicsimtorealpolicytransfer}:} Transic achieves sim-to-real transfer by leveraging human correction trajectories via a residual network architecture. It serves as a comparison method without specific geometry-aware knowledge.
\end{itemize}

As shown in Table~\ref{tab:single_task_performance}, the Direct Deploy baseline's 3.1\% average success rate confirms a substantial sim-to-real gap. The action residual method offers only marginal improvement (9.2\%), which strongly suggests that applying residuals only in the action space, without concurrently addressing the significant observation sim-to-real gap, is insufficient to overcome the reality gap.

Nevertheless, our Geo-MoE module consistently outperforms all baselines with a 50.0\% average success rate. We attribute this to its dynamic use of geometric knowledge. Unlike methods that address the gap using the overall point cloud, our approach prioritizes distinct geometric properties. This allows the model to learn more fine-grained, geometrically-aware knowledge for precise adaptation.

\begin{table}[t]
\centering
\caption{Single task sim-to-real transfer performance. 
Our Geo-MoE significantly outperforms all baselines, demonstrating the advantage of specific activated geometric knowledge in bridging the sim-to-real gap.}
\label{tab:single_task_performance}
\resizebox{1.0\linewidth}{!}{%
\begin{tabular}{@{}cccccc@{}}
\toprule
Method & Pick Cube & Stack Cube & Pick Banana & Plug Insert & Avg. SR \\
\midrule
Direct Deploy & 5.7\% & 0\% & 6.7\% & 0\% & 3.1\% \\
Action Residual & 16.7\% & 3.3\% & 13.3\% & 0.0\% & 9.2\% \\
Transic\footnotemark & 66.7\% & 30.0\% & 23.3\% & 33.3\% & 38.3\% \\
Geo-MoE & \textbf{80.0\%} & \textbf{43.3\%} & \textbf{40.0\%} & \textbf{36.7\%} & \textbf{50.0\%} \\
\bottomrule
\end{tabular}%
}
\vspace{-1em}
\end{table}

\footnotetext{We reproduced this method, adapting it to align with the distinct point cloud configuration utilized in our approach.}

\begin{table*}[t]
\vspace{-1em}
\centering
\caption{Continual learning performance across four sequential sim-to-real tasks (learned in the order: Pick Cube $\to$ Stack Cube $\to$ Pick Banana $\to$ Plug Insert). Our method (GeCo-SRT) achieves the best overall performance, with the highest average success rate and lowest average forgetting. This performance demonstrates Geo-MoE's advantage in leveraging specific geometric knowledge to bridge the sim-to-real gap, combined with Geo-PER's ability to mitigate catastrophic forgetting by continuously updating idle experts.}
\label{tab:tab_main_results_cross}
\resizebox{1.0\linewidth}{!}{%
\footnotesize 
\begin{tabular}{@{}c|c|c|c|c|c|c|c|c|c|c@{}}
\hline
\multirow{2}{*}{Method} & \multicolumn{2}{c|}{Pick Cube} & \multicolumn{2}{c|}{Stack Cube} & \multicolumn{2}{c|}{Pick Banana} & \multicolumn{2}{c|}{Plug Insert} & \multicolumn{2}{c}{Avg.} \\
\cline{2-11}
& \scriptsize $SR\uparrow$ & \scriptsize $N-NBT\downarrow$ & \scriptsize $SR\uparrow$ & \scriptsize$N-NBT\downarrow$ & \scriptsize$SR\uparrow$ & \scriptsize $N-NBT\downarrow$ & \scriptsize $SR\uparrow$ & \scriptsize $N-NBT\downarrow$ & \scriptsize $SR\uparrow$ & \scriptsize$N-NBT\downarrow$ \\
\hline

Direct Deploy & $5.7\%$ & / & $0.0\%$ & / & $6.7\%$ & / & $0.0\%$ & / & $3.1\%$ &/ \\
Naive Fine-tuning & $16.7\%$ & $100\%$ & $3.3\%$ & $100.0\%$ & $13.3\%$ & $100\%$ & $3.3\%$ & / & $9.2\%$ & $75.0\%$ \\
Transic + PER & $76.7\%$ & $81.2\%$ & $30.0\%$ & $72.3\%$ & $20.0\%$ & $66.5\%$ & $33.3\%$ & / & $40.0\%$ & $55.0\%$ \\
Geo-MoE + PER & $83.3\%$ & $34.6\%$ & $50.0\%$ & $76.8\%$ & $46.7\%$ & $7.1\%$ & $43.3\%$ & / & $55.7\%$ & $29.6\%$ \\
Geo-MoE + EWC & $80.0\%$ & $70.8\%$ & $36.7\%$ & $77.3\%$ & $20.0\%$ & $50.0\%$ & $16.7\%$ & / & $38.3\%$ & $49.5\%$ \\
Ours (GeCo-SRT) & $\textbf{86.7\%}$ & $\textbf{28.3\%}$ & $\textbf{53.3\%}$ & $\textbf{72.0\%}$ & $\textbf{60.0\%}$ & $\textbf{5.5\%}$ & $\textbf{53.3\%}$ & / & $\textbf{63.3\%}$ & $\textbf{26.5\%}$ \\
\hline
\end{tabular}%
}
\vspace{-1em}
\end{table*}
\subsubsection{Cross-Task Continual Sim-to-Real Transfer}
We further evaluate the method's ability to sequentially learn the sim-to-real gaps of distinct tasks while mitigating catastrophic forgetting. In this continual cross-task sim-to-real transfer setup, the model trains on tasks sequentially, utilizing a dataset that combines current task trajectories with replayed experiences from previous tasks

\par We compare our method against several baselines:
\begin{itemize}
    \item \textbf{Naive Fine-tuning}:  This baseline fine-tunes on the new task's data without any continual learning mechanism.
    \item \textbf{Transic + PER}: Transic combined with standard PER.
    \item \textbf{Geo-MoE + EWC}: Our Geo-MoE module combined with Elastic Weight Consolidation (EWC).
    \item \textbf{Geo-MoE + PER}: Our Geo-Moe module with standard Prioritized Experience Replay.

\end{itemize}

\par The results in Table~\ref{tab:tab_main_results_cross} show that the Naive Fine-tuning baseline suffers from catastrophic forgetting, as new task data overwrites prior knowledge.
While conventional continual learning methods like Geo-MoE + EWC and Transic + PER remain suboptimal, exhibiting significant forgetting and limited average success rates (e.g., $\leq 40.0\%$).

\par A critical insight emerges when comparing Transic + PER with Geo-MoE + PER. Simply replacing the backbone with our Geo-MoE module significantly boosts the average success rate (from 40.0\% to 55.7\%) and reduces forgetting. This strongly suggests the Geo-MoE architecture is inherently superior for continual learning. Its design, routing information via local geometric features, enables specialized experts to be activated and adapted, facilitating more effective knowledge reuse across tasks.

However, this specialized knowledge requires active protection. Our GeCo-SRT method integrates our proposed Geo-PER strategy, achieving the best-in-class performance (63.3\% average success). This refinement works because Geo-PER is tailored to the MoE structure. By shifting the prioritization metric from task-level loss to expert-level utilization, it ensures all experts—even those idle on the current task—are periodically refreshed from the buffer. This actively preserves their specialized knowledge and leads to robust retention.

In summary, our GeCo-SRT method combines a superior architecture for knowledge reuse (Geo-MoE) with a targeted strategy for knowledge protection (Geo-PER), achieving the most robust continual sim-to-real performance.

\begin{table*}[t]
\vspace{-1em}
\centering
\caption{Ablation study on core components of Geo-MoE. For the Observation Residual component, $\checkmark$ indicates using our point cloud encoder to align the observation sim-to-real gap, whereas $\times$ denotes using the action residual module. For the MoE component, $\checkmark$ indicates using our mixture-of-experts architecture to obtain distinct knowledge, while $\times$ represents the baseline without it.
}
\label{tab:ablation} 
\resizebox{1.0\linewidth}{!}{
\footnotesize 
\begin{tabular}{@{}cc|c|c|c|c|c|c|c|c|c|c@{}} 
\hline
\multirow{2}{*}{\makecell{ Observation \\ Residual} } & \multirow{2}{*}{{\makecell{ MoE}}} & \multicolumn{2}{c|}{Pick Cube} & \multicolumn{2}{c|}{Stack Cube} & \multicolumn{2}{c|}{Pick Banana} & \multicolumn{2}{c|}{Plug Insert} & \multicolumn{2}{c}{Avg.} \\
\cline{3-12} 

& & \scriptsize $SR\uparrow$ & \scriptsize $N-NBT\downarrow$ & \scriptsize $SR\uparrow$ & \scriptsize$N-NBT\downarrow$ & \scriptsize$SR\uparrow$ & \scriptsize $N-NBT\downarrow$ & \scriptsize $SR\uparrow$ & \scriptsize $N-NBT\downarrow$ & \scriptsize $SR\uparrow$ & \scriptsize$N-NBT\downarrow$ \\ 

\hline

$\times$ &$\times$ &$6.7\%\text{ } $&$ 100\% $&$3.3\% $&$100.0\% $&$3.3\% $&$100.0\%  $&$0.0\% $&/&$3.3\%  $&$75.0\% $ \\ 
$\times$&$\checkmark$ &$16.7\% $&$86.7\% $ &$3.3\% $ &$100.0\%  $&$13.3\% $ &$75.2\%  $&$3.3\% $ &/ & $9.2\% $&$65.5\%  $\\ 
$\checkmark$&$\times$ &$\textbf{83.3\%}  $&$37.3\%  $&$46.7\%  $&$96.5\% \text{ } $&$23.3\%  $&$14.2\%  $&$30.0\% $ &/ &$45.8\% $ &$37.0\%  $\\ 
$\checkmark$&$\checkmark$ &$\textbf{83.3\%} $ &$\textbf{34.6\%}  $&$\textbf{50.0\%}  $&$\textbf{76.8\%}\text{ } $&$\textbf{46.7\%}  $&$ \textbf{7.1\%} $&$ \textbf{43.3\%} $&/ &$\textbf{55.8\%} $&$\textbf{29.6\%}  $ \\ 

\hline
\end{tabular}%
\vspace{-1em}
}

\end{table*}

\subsubsection{Cross-Task Transfer Efficiency}
\begin{figure}[t]
    \centering
    \includegraphics[width=1.0\linewidth]{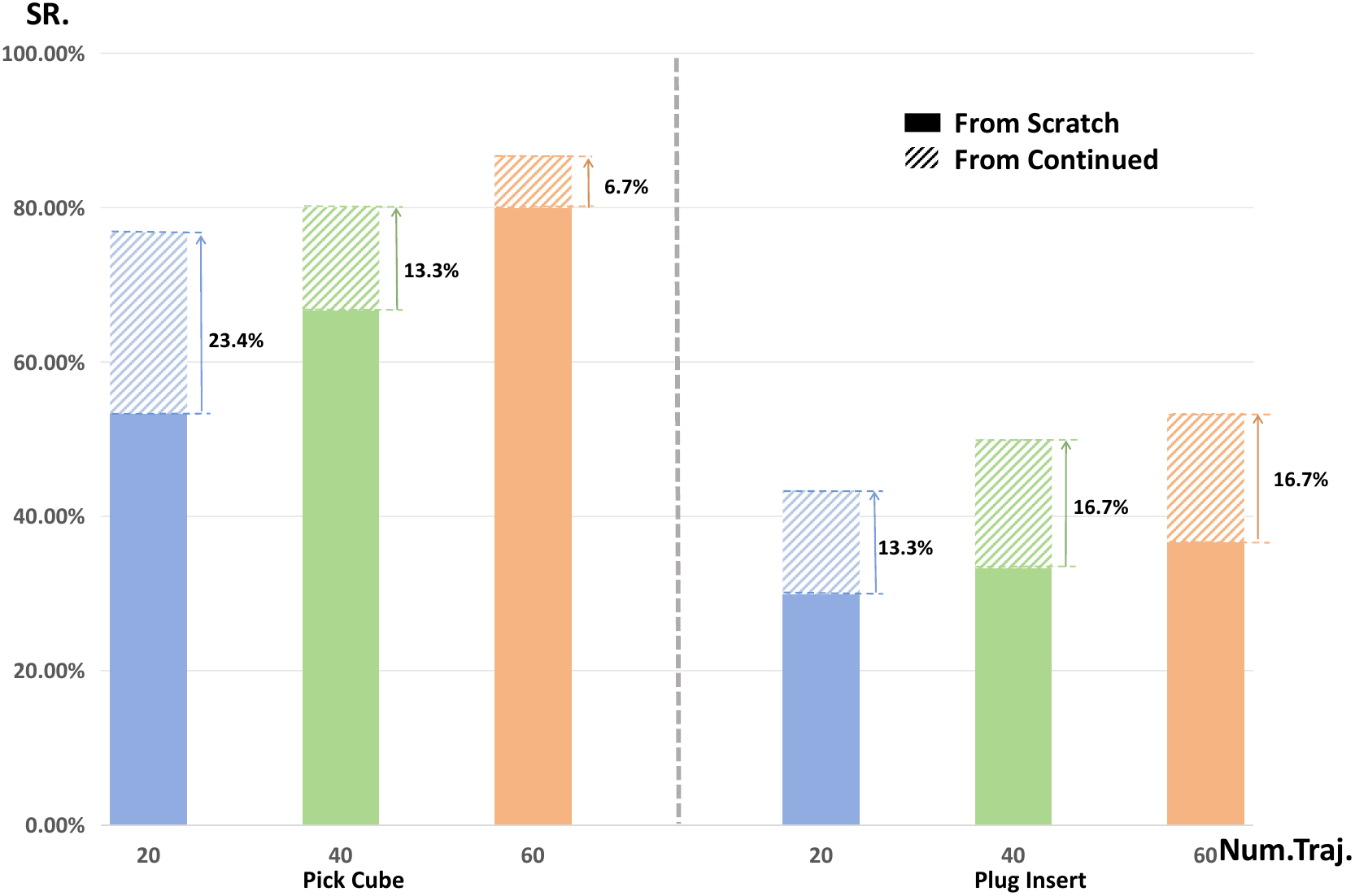}
    \caption{Data efficiency comparison of training ``From Scratch'' vs. ``From Continued''. The performance gain (hatched area) highlights our method's ability to utilize transferable geometric knowledge for cross-task sim-to-real transfer, which enhances data efficiency when adapting to new tasks, especially in low-data regimes.}
    \label{fig:sample_efficiency}
    \vspace{-1em}
\end{figure}

We evaluate sample efficiency in Figure~\ref{fig:sample_efficiency}. We compare two training scenarios: training from scratch (using only the new task's data) and our continual learning approach (transferring from a pre-trained task). Both scenarios are evaluated using 20, 40, and 60 correction trajectories.

The results clearly illustrate the significant benefit of our method. For the Pick Cube task, the model trained via continual learning achieves 76.6\% success with only 20 trajectories. This performance dramatically surpasses the from-scratch baseline at the same data point and nearly matches the performance the from-scratch model achieves with $3 \times$ the data (60 trajectories). This trend is consistent for the Plug Insert task, where our method with 20 trajectories achieves a performance level that the from-scratch model requires more than 60 trajectories to reach.

This experiment confirms that our continual learning method significantly improves sample efficiency. The agent's ability to adapt so rapidly is not merely due to parameter initialization; it demonstrates that our method successfully learns and reuses transferable geometric knowledge about the sim-to-real gap. By leveraging this shared cross-task knowledge, the agent can adapt to new tasks with limited correction demonstrations, which is critical for real-world applications.

\subsection{Ablation Results} \label{sec:ablation}
Table~\ref{tab:ablation} highlights each component's contribution.
The baseline model (Observation Residual $\times$, MoE $\times$) performs poorly, and sees negligible improvement when only the MoE module is added ($\times$, $\checkmark$). This suggests that the expert architecture is ineffective without informative features to bridge the observation sim-to-real gap.

The most significant gain comes from introducing our Observation Residual component ($\checkmark$, $\times$). This change alone causes the average success rate to leap to 45.8\% and average N-NBT to plummet to 37.0\%. This provides a stable feature bridge for visual alignment, effectively mitigating the sim-to-real gap.

Building on this strong Observation Residual component, adding the MoE module ($\checkmark$, $\checkmark$) achieves the best performance. This highlights the MoE's synergistic role: it takes the stable geometric features and enables specialized handling. By routing geometric patterns to distinct experts, the model processes local discrepancies in a more targeted manner. This specialization improves performance on later tasks and further mitigates forgetting.

In summary, the ablation confirms our components are complementary: the Observation Residual component extracts a stable, transferable feature representation based on sim-real invariant geometry, while the MoE module builds on this to specialize adaptation and preserve knowledge.

\subsection{Cross-task Transfer Results} \label{sec:cross_transfer}

\par Finally, we investigate how task similarity impacts the effectiveness of knowledge transfer, particularly in a data-scarce scenario (using only 10 correction trajectories). We compare training ``From scratch'' versus transferring from various pre-trained source tasks, with results in Table~\ref{tab:cross_task}.

The results indicate that task geometric similarity is a critical factor for our geometry-aware continual adaptation method. This is evidenced by significant positive transfer when learning the Stack Cube task after the similar Pick Cube task, far outperforming the baseline. Conversely, negative transfer occurs when transferring from the dissimilar PlugInsert task. Notably, the transfer from PickBanana to PlugInsert is also beneficial, which we attribute to the shared geometric challenge of grasping non-cuboid objects.

While the primary insight concerns task similarity, this positive transfer is also highly data-efficient. For instance, the success rate on Stack Cube with only 10 trajectories nearly matches the performance of the from-scratch model trained with 60 trajectories.

This experiment confirms our geometry-aware method learns transferable geometric knowledge. The model's ability to reuse knowledge from geometrically similar tasks, while failing with dissimilar ones, demonstrates it captures and leverages foundational geometric understanding for rapid, low-data adaptation.

\begin{table}[H]
  \centering
  \caption{Impact of task similarity on knowledge transfer. All models use 10 trajectories unless otherwise specified.}
  \label{tab:cross_task}
  \begin{tabular}{c|c|c}
    \hline 
    \textbf{Evaluate Task} & \textbf{Pre-trained Task} & \textbf{SR} \\
    \hline
    \multirow{5}{*}{StackCube}
    & PickCube & 40.0\% \\
    & PickBanana & 33.3\% \\
    & PlugInsert & 16.7\% \\
    & From scratch & 26.7\% \\
    \cline{2-3} 
    &$\text{From scratch (60 traj.)}$ & 43.3\% \\
    \hline 
    \multirow{5}{*}{PlugInsert}
    & PickCube & 33.3\% \\
    & StackCube & 40.0\% \\
    & PickBanana & 30.0\% \\
    & From scratch & 23.3\% \\
    \cline{2-3} 
    &$\text{ From scratch (60 traj.)}$ & 36.7\% \\
    \hline 
  \end{tabular}
  \vspace{-1em}
\end{table}

\section{Conclusion}
\label{sec:conclution_and_limitation}
In this work, we build a continual cross-task sim-to-real transfer paradigm for knowledge accumulation across iterative transfers, enabling effective and efficient adaptation to novel tasks.
To achieve this, we propose the GeCo-SRT, a geometry-aware continual adaptation method. It extracts domain-invariant and task-invariant knowledge from local geometric features to accelerate subsequent adaptation. This method consists of a geometry-aware mixture-of-experts module to obtain specialized transferable knowledge, and a geometry-expert-guided prioritized experience replay strategy to mitigates catastrophic forgetting with the guidance of expert utilization. Results show a 52\% average performance improvement over the baseline and demonstrate significant data efficiency, adapting to new tasks with only 16.7\% data. We hope our work provides insight into continual sim-to-real transfer by effectively leveraging local geometric knowledge.

\textbf{Discussion and Future Work.} 
While GeCo-SRT has demonstrated effectiveness and efficiency in cross-task sim-to-real transfer with continual knowledge accumulation, our method primarily focuses on bridging the observation gap by leveraging local geometric features.
This specific focus may limit its applicability in non-geometric sim-to-real gaps (e.g., complex dynamics). Furthermore, we believe exploring multiple modalities, such as the semantic relationships between tasks, could provide a promising avenue for effective cross-task sim-to-real transfer.

\section{Acknowledgement}
This work is supported by Public Computing Cloud, Renmin University of China, and fund for building world-class universities (disciplines)
of Renmin University of China.
{
    \small
    \bibliographystyle{ieeenat_fullname}
    \bibliography{main}
}

\maketitlesupplementary
\section{Implementation and Experimental Analysis}
\label{sec:supp}
In this section, we provide detailed experimental settings, including simulation training, policy learning with 3D representation, task descriptions, human-in-the-loop data collection, and a qualitative analysis of action adaptation.

\subsection{Simulation Training}
We use ManiSkill (v3.0.0b21) as the simulator backend. ManiSkill is built on the SAPIEN engine, which utilizes NVIDIA PhysX 5 as the physics engine to provide realistic and precise simulation. The robot models and task environments are adapted from the official ManiSkill benchmark suite, which provides a diverse set of manipulation tasks involving a variety of object types and geometric structures.

\subsection{Policy Learning with 3D Representation}  Typical RGB observation used in visuomotor policy training suffers from several drawbacks that hinder successful transfer, such as vulnerability to different camera poses and discrepancies between synthetic and real images. To bypass these issues, we propose to use point cloud as the main visual modality. 3D point clouds make it easier for the policy to learn and generalize the geometric knowledge of different objects.

\par During the simulation training phase, we set up $N=2$ dual cameras to capture RGBD images. For each camera view, we obtain the raw point cloud $P_{\text{cam}}^{(i)} \in \mathbb{R}^{K \times 3}$ in its local camera frame. We then transform it into the robot's base frame using the camera's known position and orientation:
\begin{equation}
P_{\text{base}}^{(i)} = P_{\text{cam}}^{(i)} (R^{(i)})^T + (p^{(i)})^T 
\end{equation}
Here, $R^{(i)} \in \mathbb{R}^{3 \times 3}$ and $p^{(i)} \in \mathbb{R}^{3 \times 1}$ denote the $i$-th camera's orientation and translation in the base frame, respectively. This process is identical to the method used on the real robot (which uses camera calibration), thus aligning the point cloud acquisition method to reduce the sim-to-real gap. After transforming both views to the base coordinate system, the complete scene point cloud $P^S$ is aggregated by concatenating the views:

\begin{equation}
     P^S = \bigcup_{i=1}^{N} P_{\text{base}}^{(i)} 
\end{equation}
This aggregated point cloud is then downsampled via Farthest Point Sampling (FPS) and subsequently used as the policy network input.
\begin{figure}[t]
  \centering
  \includegraphics[width=\linewidth]{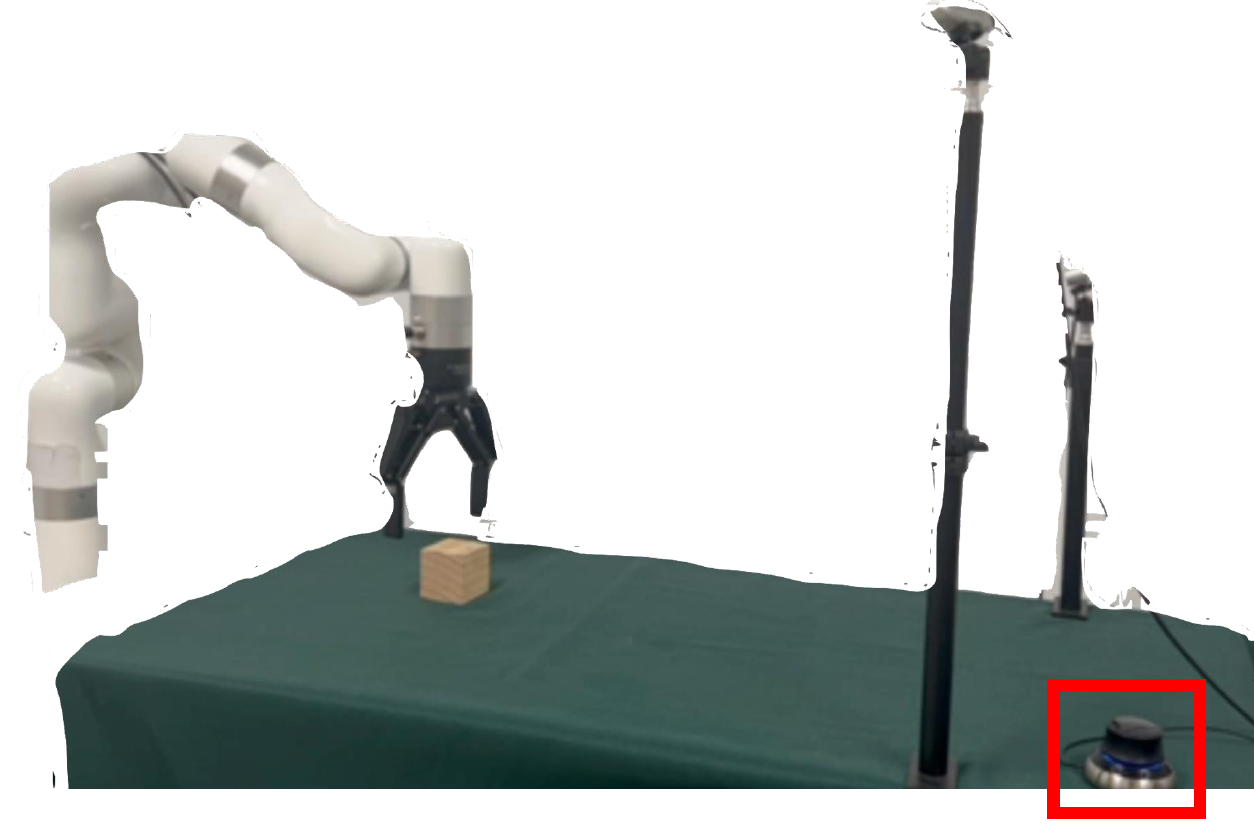}
  \caption{Real-world workspace setup for human-in-the-loop data collection. The human operator provides online correction through a 3Dconnexion SpaceMouse (marked with a red box) while monitoring the robot’s execution.}
  \label{fig:setup_supp}
  \vspace{-1em}
\end{figure}

\begin{figure*}[t]
 \vspace{-1em}
 \centering
 \includegraphics[width=\linewidth]{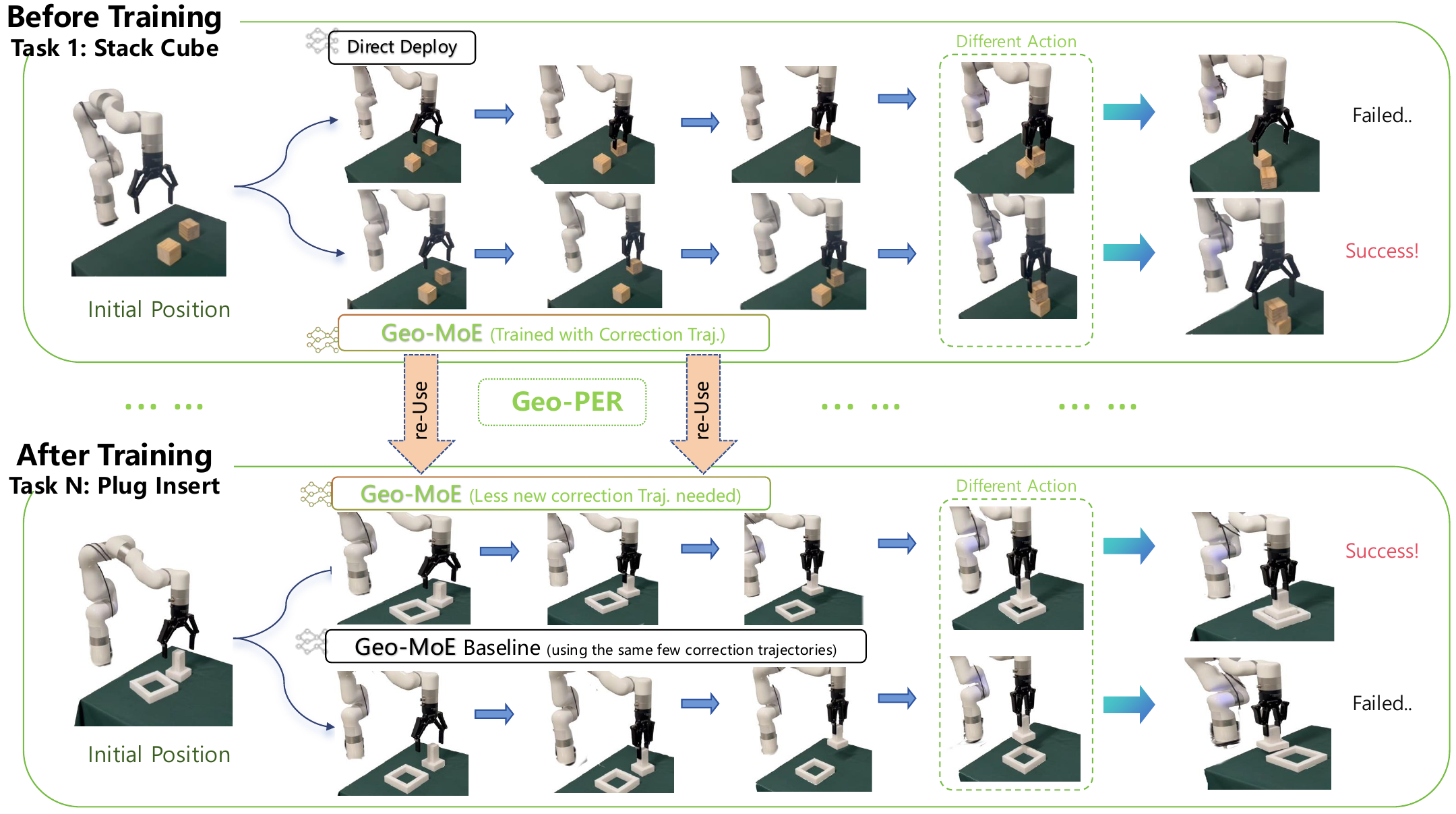}
 \caption{Qualitative visualization of action adaptation on the ``Stack Cube'' and ``Plug Insert'' task. \textbf{(Top)} In the initial phase (Task 1), human correction trajectories are used to rectify the failure modes of direct deployment. \textbf{(Bottom)} In the continual learning phase (Task N), our Geo-MoE model utilizes the \textit{Re-Use} mechanism to achieve success with significantly less new correction data compared to the baseline, demonstrating high data efficiency.}
\label{fig:action_comparison}
 \vspace{-1em}
\end{figure*}

\subsection{Task Description}
\par To verify module transferability, we designed four manipulation tasks of varying difficulty: Pick Cube, Stack Cube, Pick Banana, and Plug Insert. For all tasks, the robot gripper starts at a predefined fixed location, with the tabletop serving as the origin plane. We detail each task's objective, randomized initial conditions, and success criteria below.
\begin{itemize}
    \item \textbf{Pick Cube:} The robot must grasp and lift a cube. The cube is initialized at a random position within a specified range on the tabletop, and the task is successful once the cube is grasped and lifted.

    \item \textbf{Stack Cube:} The robot must grasp the left of two parallel cubes and stack it onto the right one. The two cubes are initialized in parallel with a set gap at random positions. Success requires the target cube to be stably placed on the other cube without falling.

    \item \textbf{Pick Banana:} The robot must grasp and lift a banana-shaped object. The object is initialized at a random position, but its orientation is fixed perpendicular to the gripper's opening. Success is defined as lifting the object without gripping the tablecloth or causing gripper deformation.
    
    \item \textbf{Plug Insert:} The robot must pick up a plug and insert it into a socket. The plug and socket are initialized at random positions, with the plug always to the left of the socket. Success requires the plug to be steadily inserted into the socket, without gripping the tablecloth or causing gripper deformation.
\end{itemize}

\subsection{Human Correction Data Collection}

\par We implement a human-in-the-loop mechanism to harvest high-quality Human Correction Data, and our data collection workspace is shown at \ref{fig:setup_supp}. The base policy is first deployed to execute the task autonomously. When a policy failure or deviation is observed, the human operator intervenes using a 3Dconnexion SpaceMouse, taking control of the loop to ensure the completion of a successful trajectory. To guaranty the integrity of the training data, a post-processing step is applied: segments corresponding to the policy's erroneous actions prior to the intervention are pruned. This selective filtering ensures that the model updates are based solely on high-quality expert demonstrations, avoiding the negative transfer of failure modes.

\subsection{Qualitative Analysis of Action Adaptation}
To intuitively understand how our method improves performance, we visualize the action execution trajectories in Figure~\ref{fig:action_comparison}.

\noindent\textbf{Correction for the Sim-to-Real Gap (Task 1).} 
As shown in the top row of Figure~\ref{fig:action_comparison}, the direct deploy policy suffers from domain gaps, generating deviant actions (labeled as ``Different Action'') that lead to task failure. By introducing human-in-the-loop intervention, our module leverages distinct geometric cues to bridge this gap. This capability allows the model to learn a fine-grained spatial understanding, ensuring the precise action rectification required to successfully complete the ``Stack Cube'' task.

\noindent\textbf{Efficiency via Knowledge Re-use (Task N).} 
The core advantage is highlighted in the bottom row. Driven by the synergy between Geo-PER and Geo-MoE, our framework transcends simple parameter initialization to actively reuse transferable geometric knowledge. This shared cross-task understanding enables the agent to adapt to Task N with minimal correction data, whereas the baseline fails under the same conditions. This comparison confirms that GeCo-SRT significantly enhances data efficiency by effectively leveraging geometric priors.

\subsection{Scalability and Autonomy of HITL}
By selectively addressing critical failures, our Human-in-the-Loop (HITL) framework achieves significant performance gains in long-horizon tasks with fewer than 50 human interventions, demonstrating exceptional efficiency and scalability. Furthermore, the framework is agnostic to the correction source; it can seamlessly integrate with MLLM-based agents~\cite{xia2025robotic} to facilitate autonomous online error recovery. This transition from human-in-the-loop to model-in-the-loop eliminates manual dependency and further broadens the applicability of our approach in complex, autonomous environments.

\begin{table}[b]
    \centering
    \caption{\textbf{Sensitivity Analysis of Expert Numbers ($N$).} We evaluate the impact of $N$ on Success Rate (SR) and Net Benefit of Transfer (NBT). Performance remains robust across configurations, with $N=3$ providing an optimal balance between efficiency and specialization.}
    \label{tab:moe_sens}
    \begin{tabular}{ccc}
        \toprule
        Number of Experts ($N$) & Avg. SR (\%) $\uparrow$ & NBT (\%) $\downarrow$ \\ 
        \midrule
        2 & 60.0 & 40.0 \\
        3 & \textbf{66.7} & 33.3 \\
        8 & 65.0 & \textbf{30.0} \\
        \bottomrule
    \end{tabular}
\end{table}

\subsection{Computational Efficiency and Robustness}
Our lightweight point-cloud residual network achieves a real-time inference latency of 26.1ms, competitive with state-of-the-art baselines such as Transic (25.6ms) and Direct Deploy (19.1ms). Our method also exhibits high robustness to hyperparameter variations, particularly regarding the number of experts (Tab.~\ref{tab:moe_sens}). 

In terms of memory management, a fixed-size replay buffer retaining 50\% of historical data maintains performance parity with full-history training. Specifically, in the Pick-to-Stack Cube task transfer, the N-NBT forgetting rate is 20.0\% with a full buffer and only 23.3\% with a 50\% buffer. When coupled with Geo-PER, this strategy effectively constrains memory growth while ensuring robust knowledge retention.

\begin{figure}[t]
 \centering
 \includegraphics[width=\linewidth]{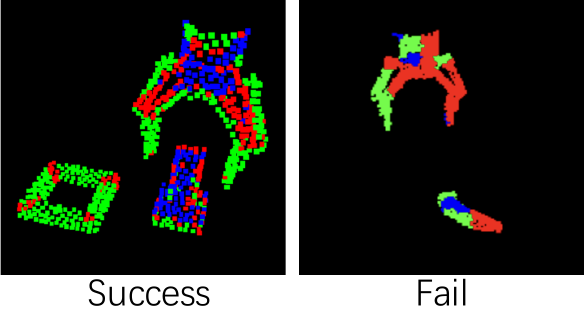}
 \caption{\textbf{Qualitative Visualization of Expert Specialization.} Different colors indicate the routing preference of each expert. Our MoE layer naturally learns to partition the input space based on geometric primitives, specializing in edges, corners, and planar surfaces.}
\label{fig:fig2}
\end{figure}

\subsection{MoE Interpretability and Failure Modes}
As illustrated in Fig.~\ref{fig:fig2}, the routing mechanism exhibits clear interpretability: experts specialize in distinct geometric features. We identify \textit{routing collapse} as a primary failure mode under challenging Out-of-Distribution (OOD) observations (e.g., unseen geometries). In these cases, the gating network disproportionately routes most points to only 1--2 experts, which strongly correlates with subsequent contact failures or grasping instabilities.

\begin{table}[ht]
    \centering
    \caption{\textbf{Quantitative Results on New Tasks.} Success Rate (SR) for Faucet and Tidying tasks. Our Geo-MoE (continual) significantly outperforms zero-shot and scratch-trained baselines.}
    \label{tab:real_world_results}
    \begin{tabular}{lcc}
        \toprule
        Method & Faucet (\%) $\uparrow$ & Tidying (\%) $\uparrow$ \\
        \midrule
        Direct Deploy & 10.0 & 0.0 \\
        Geo-MoE (zero-shot) & 53.3 & 30.0 \\
        \midrule
        Geo-MoE (scratch) & 76.6 & 43.3 \\
        \textbf{Geo-MoE (continual)} & \textbf{83.3} & \textbf{56.7} \\
        \bottomrule
    \end{tabular}
\end{table}

\begin{figure}[ht]
    \centering
    \includegraphics[width=\linewidth]{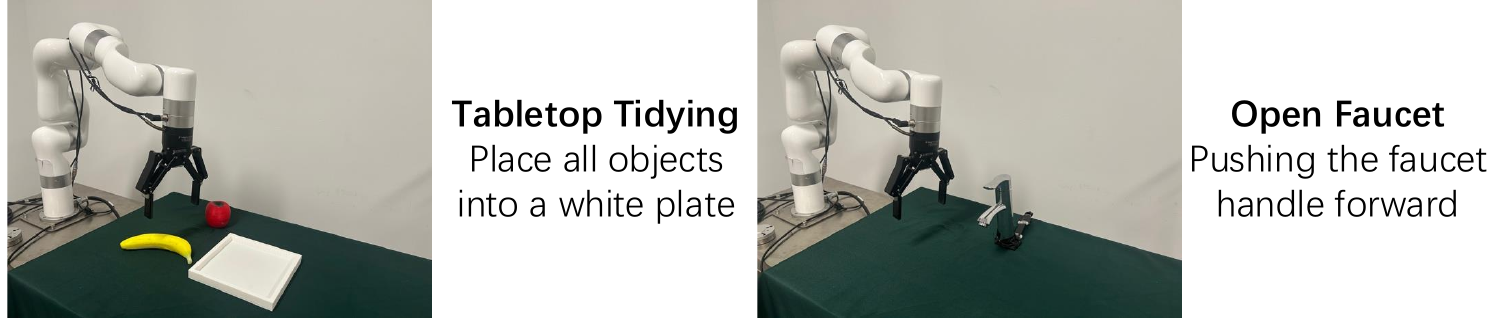} 
    \caption{\textbf{Visualization of Newly Integrated Tasks.} Experimental setups for \textit{Faucet} and \textit{Tidying} scenarios. Our method effectively handles these novel geometries through efficient knowledge transfer.}
    \label{fig:fig4}
\end{figure}

\subsection{Task Complexity and Diversity}
\begin{figure}[ht]
    \centering
    \includegraphics[width=\linewidth]{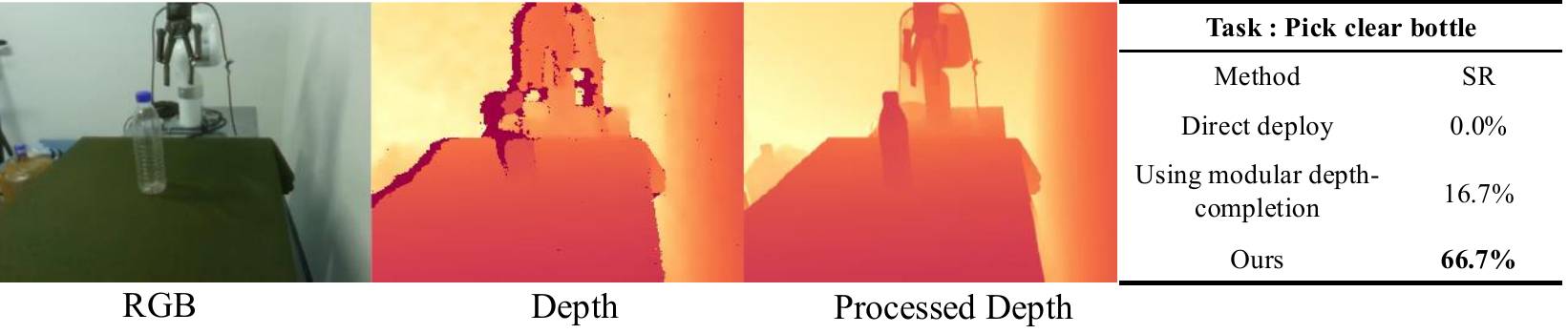}  
    \caption{\textbf{Qualitative Results of Depth Completion.} Our method effectively restores missing depth information in geometrically complex regions, providing a dense, noise-resilient point cloud for downstream manipulation.}
    \label{fig:fig1}
\end{figure}

To further evaluate the diversity of our approach, we introduce three challenging scenarios: \textit{Open Faucet} (nonlinear rotation), \textit{Tabletop Tidying} (long-horizon), and \textit{Pick Clear Bottle} (transparency) (Figs.~\ref{fig:fig4}, \ref{fig:fig1}). Tab.~\ref{tab:real_world_results} compares our method against Direct Deploy. Pretrained on the four tasks in the main text, our Geo-MoE module achieves strong zero-shot performance without real-world interaction data. Furthermore, the continual learning setting significantly yields higher success rates than training from scratch, confirming that GeCo-SRT effectively accumulates transferable knowledge for data-efficient adaptation.

\subsection{Handling Depth Invalidity} 
Our core contribution models sim-to-real as a continual learning problem anchored on geometric features; sensor-level limitations are orthogonal to this framework. However, to demonstrate extensibility, we integrate a modular depth-completion step~\cite{liu2025manipulation}. By recovering metric depth from RGB-D inputs (Fig.~\ref{fig:fig1}), we ensure sufficient geometric fidelity, enabling Geo-MoE to maintain robust transfer even on challenging surfaces like clear bottles. 

\subsection{Evaluation with RGB Inputs} 
We further evaluate GeCo-SRT using RGB-only inputs (Tab.~\ref{tab:rgb}). While RGB introduces larger sim-to-real gaps, our geometry-aware experts operate on image patches to achieve reasonable performance. This demonstrates that our framework generalizes to other modalities, although point clouds remain superior due to their direct representation of the geometric invariance that motivates our design.

\begin{table}[ht]
    \centering
    \caption{\textbf{Success Rate (SR) with RGB Input.} Comparison under RGB-only observations. $T1^{\dagger}$ and $T2^{\dagger}$ denote test scenarios with increased geometric complexity. $^{\ast}$ indicates results under partial observation.}
    \label{tab:rgb}
    \begin{tabular}{lcc}
        \toprule
        Method & $T1^{\dagger}$ (\%) $\uparrow$ & $T2^{\dagger}$ (\%) $\uparrow$ \\ 
        \midrule
        Direct Deploy         & 3.3           & 0.0 \\
        Domain Randomization  & 10.0          & 6.7 \\
        \textbf{Ours}         & \textbf{40.0} & \textbf{20/33}$^{\ast}$ \\ 
        \bottomrule
    \end{tabular}
\end{table}

\end{document}